\documentclass[runningheads]{llncs}
\usepackage[T1]{fontenc}
\usepackage{graphicx}
\usepackage{amssymb}
\usepackage{amsmath}
\usepackage{xcolor}
\usepackage{float}
\usepackage{tabularx}
\usepackage[titletoc,toc,title]{appendix}
\usepackage{fontawesome5}   % newer version of fontawesome
\usepackage{hyperref}
\begin{document}

\title{A public cardiac CT dataset featuring the left atrial appendage}

\author{\textsuperscript{1}Bjørn Hansen, \textsuperscript{2}Jonas Pedersen, \textsuperscript{2}Klaus F. Kofoed, \textsuperscript{3}Oscar Camara, \textsuperscript{1}Rasmus R. Paulsen and \textsuperscript{1}\textsuperscript{,4}Kristine Sørensen}
\institute{\textsuperscript{1}DTU Compute, Technical University of Denmark, Denmark \\
\textsuperscript{2}Copenhagen University Hospital Rigshospitalet, Denmark\\
\textsuperscript{3}Universitat Pompeu Fabra, Spain\\
\textsuperscript{4}Novo Nordisk, Denmark\\
    \email{bmsha@dtu.dk}}
\maketitle              % typeset the header of the contribution
\begin{abstract}
Despite the success of advanced segmentation frameworks such as TotalSegmentator (TS), accurate segmentations of the left atrial appendage (LAA), coronary arteries (CAs), and pulmonary veins (PVs) remains a significant challenge in medical imaging. In this work, we present the first open-source, anatomically coherent dataset of curated, high-resolution segmentations for these structures, supplemented with whole-heart labels produced by TS on the publicly available ImageCAS dataset consisting of 1000 cardiac computed tomography angiography (CCTA) scans. One purpose of the data set is to foster novel approaches to the analysis of LAA morphology.

LAA segmentations on ImageCAS were generated using a state-of-the-art segmentation framework developed specifically for high resolution LAA segmentation. We trained the network on a large private dataset with manual annotations provided by medical readers guided by a trained cardiologist and transferred the model to ImageCAS data. CA labels were improved from the original ImageCAS annotations, while PV segmentations were refined from TS outputs. In addition, we provide a list of scans from ImageCAS that contains common data flaws such as step artefacts, LAAs extending beyond the scanner's field of view, and other types of data defects\footnote{Data and model weights available at https://github.com/Bjonze/Public-Cardiac-CT-Dataset \href{https://github.com/Bjonze/Public-Cardiac-CT-Dataset}{\faGithub}}.

\keywords{Cardiac Computed Tomography  \and Whole heart segmentation \and labelmaps.}
\end{abstract}
\section{Introduction}
Increased life expectancy has led to a rise in heart-related events, including strokes, which account for $\sim$10\% of global death~\cite{WHO2021}.
When investigating stroke, the search for a cause often leads to the heart. Although the connection between the heart and the brain in stroke etiology is not immediately obvious, it can be explained by their shared anatomy. Oxygenated blood is pumped from the left ventricle into the aorta, and from there 15-20\% of the blood flows through the carotid and vertebral ateries to the brain. Consequently, if a blood clot forms or dislodges in either of the left heart chambers, it has a considerable risk of traveling to the brain where it may cause a stroke.  

\begin{figure}[t]
    \centering
    \includegraphics[width=\linewidth]{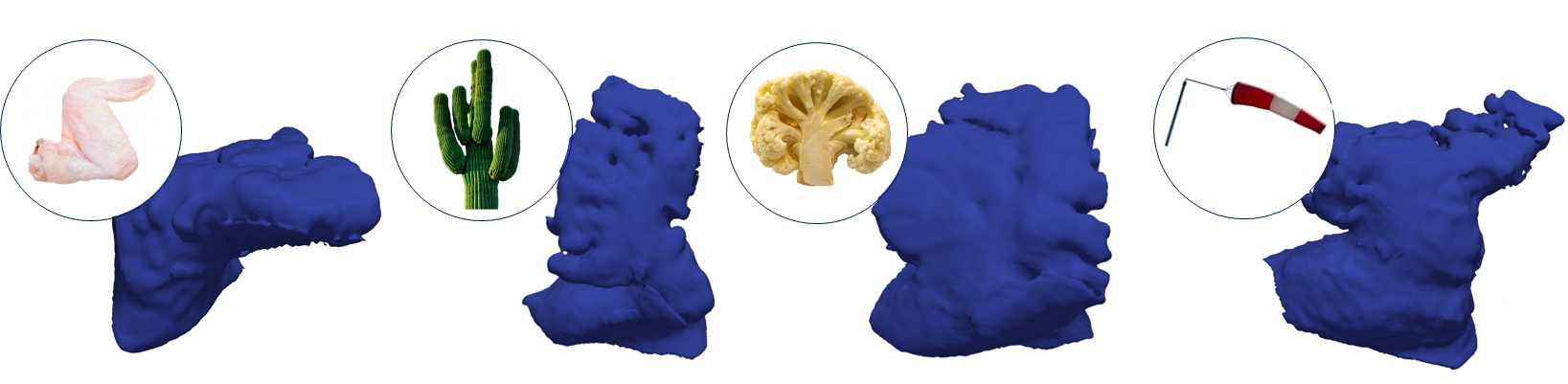}
    \caption{\textbf{Left atrial appendage morphologies.} Examples of a chicken wing, cactus, cauliflower and windsock morphology from ImageCAS following the classification system of \cite{DIBIASE2012531}.}
    \label{fig:laa_morph_fig}
\end{figure}

Studies such as \cite{blackshear1996appendage} and \cite{DIBIASE2012531} suggest that the left atrial appendage (LAA), a small finger-like extension of the left atrium (LA), may play a significant role in blood clot formation. The LAA’s size and shape vary greatly between individuals, leading to hypotheses about links between morphology and stroke risk. Clinically, LAA shapes are grouped into four categories based on resemblance: chicken wing, cactus, cauliflower, or windsock~\cite{DIBIASE2012531} (see \textbf{Fig.~\ref{fig:laa_morph_fig}}).
Several studies have investigated links between LAA shape categories and stroke risk~\cite{khurram2013,DIBIASE2012531,lupercio2016,wang2020}, but the conclussions remain inconsistent. This is partly due to low reproducibility in manual classification: Three annotators agree on the morphology in only 29\% of cases~\cite{wu2019}. Data-driven methods using statistical shape analysis have been used to study anatomy~\cite{slipsager2018statistical}. Computational fluid dynamics (CFD) modeling offers an alternative by simulating blood flow from the pulmonary veins into the LA and LAA, though results rely heavily on segmentation quality~\cite{masci2018,garc2018}.

Most studies on LAA morphology use private, manually annotated cardiac computed tomography angiography (CCTA) or Magnetic Resonance Imaging (MRI) datasets, which are typically small and not publicly available, making it difficult to share knowledge and compare methods.
Access to a larger public LAA data set will be useful for both simulating blood flow in real anatomies as well as an important enabler for developing advanced data-driven shape analysis tools to describe the complex LAA anatomy in a quantitative way. 
As blood flow and thrombus risk depends on more than LAA morphology, our data set also includes heart chambers, pulmonary veins (PV), and coronary arteries (CA), allowing research on structural relationships such as PV configuration relative to LAA placement~\cite{mill2024role}.

State-of-the-art (SOTA) methods for LAA segmentation fall into two main categories: heuristics-based and model-based approaches. Heuristic methods typically require user-selected seed points \cite{vdovjak2019,leventic2019} or a centerline \cite{Morais2018}, achieving Dice scores between 82.5\% and 91.54\% on small datasets (17–20 cases). Model-based methods, such as atlas-based segmentation \cite{Qiao2019}, report Dice scores of up to 91\% using 10 training and 20 testing MRI scans. More recent approaches employ fully convolutional networks, including 2D CNNs \cite{takayashiki2019,habijan2022} and 3D coherence as post-processing \cite{jin2018}, with Dice scores reaching 94.76\%. \cite{NUDF} introduced a fully automated, high-detail LAA segmentation model that achieves SOTA on symmetric Chamfer distance and mesh accuracy, beating previous 3D convolutional methods. In our work, we extend this model by training it on a much larger dataset of 980 CCTA images, compared to the original 76.

\begin{figure}[htbp]
\includegraphics[width=\textwidth]{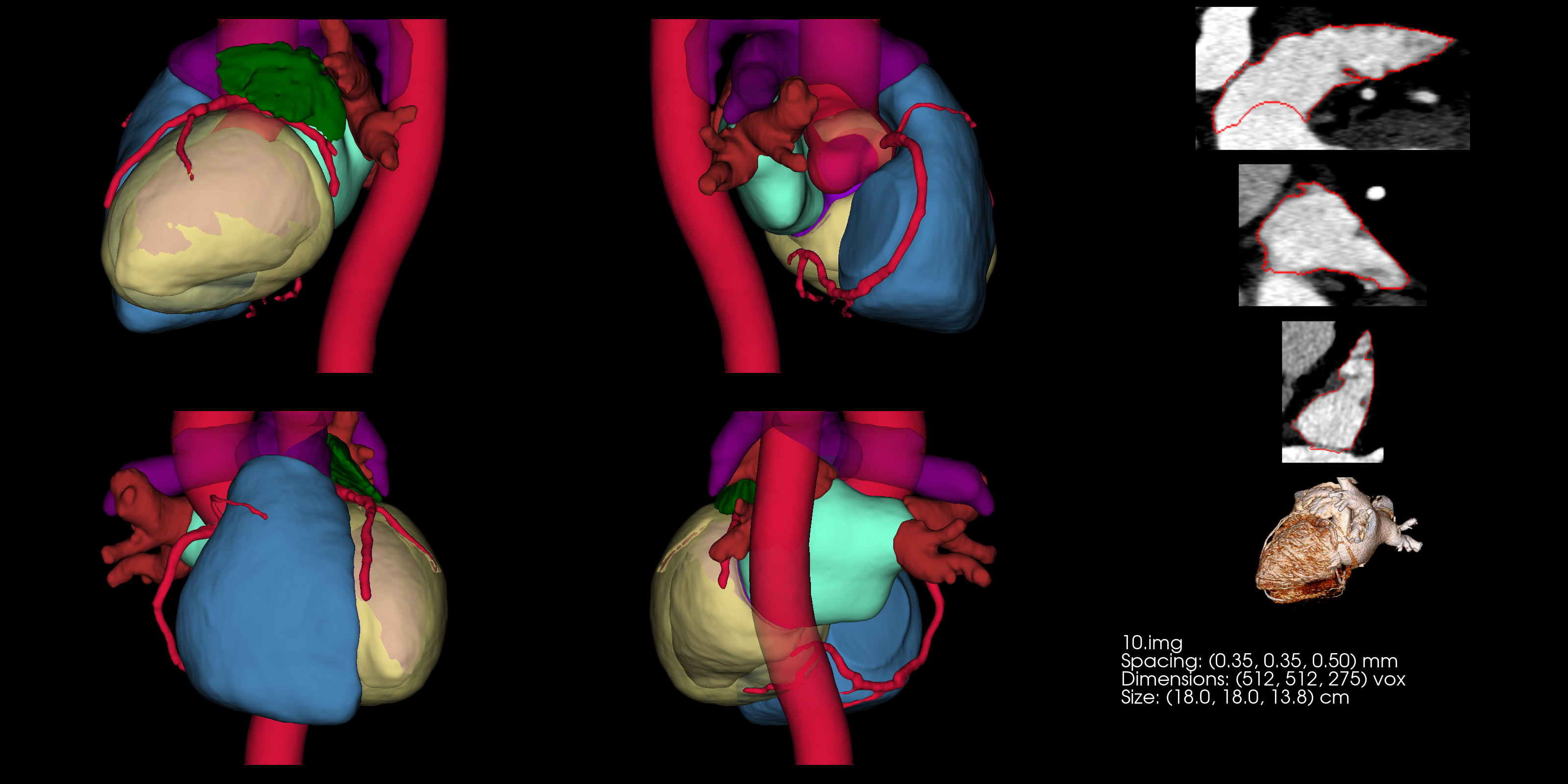}
\caption{\textbf{Example of the provided data.} Surfaces represent the fused labels, while orthogonal cross-sections of the LAA and a volume rendering are used for visual quality control. Note that in this example, the LAA (green) and the PVs (dark-red) nearly overlap.}\label{fig:VisualResult}
\end{figure}

The main contribution of this paper is the first open-source CCTA dataset with carefully curated LAA segmentation labels and complete with anatomical correct segmentations of the remaining heart structures.
These labels are build upon the existing ImageCAS dataset~\cite{ImageCASPaper}, which consists of 1000 CCTA scans with annotations of the coronary arteries. 
More specifically the papers presents the following contributions: 

\begin{itemize}
    \item LAA labels generated using a state-of-the-art segmentation network~\cite{NUDF} trained on 980 manually annotated cases from a private dataset.
    
    \item Heart structure labels (chambers, myocardium, aorta, PAs) produced by TotalSegmentator~\cite{wasserthal2023totalsegmentator}.

    \item Modified coronary artery labels based on the original dataset.

    \item All labels refined for anatomical consistency.

    \item A list of scan IDs highlighting partial structures, artefacts, or other issues.
    
\end{itemize}
Due to data regulations, only the public dataset is released, with refined labelmaps (.nii) (see \textbf{Fig.~\ref{fig:VisualResult}}).\footnote{Please note that the heart chamber labels produced by TotalSegmentator \texttt{heartchambers\_highres} is governed by a license agreement found on the TotalSegmentator Github page. It is not allowed to directly retrain alternative segmentation frameworks on this output. The other labels are not governed by this agreement.}. We hope this data will facilitate the development of novel LAA shape descriptors, CFD simulations and other research avenues requiring high quality, whole-heart segmentation data. 

\section{Data}
\textbf{Private LAA training Data.}
The LAA segmentation network framework is trained on CCTA images from the \textit{Copenhagen General Population Study}.  

The images were taken in the end diastolic phase and consist of $512 \times 512 \times (512-640)$ voxels and a spacing of 0.429mm $\times$ 0.429mm $\times$ 0.250mm. 
A subset of 980 participants had the LAA manually annotated by an expert medical annotator instructed by a trained cardiologist. The annotation was done
semi-automatically with a custom segmentation tool in \texttt{Canon Vitrea Advanced Visualization Workstation}. 

\textbf{Public labeling data.}
To make the learnings from the private annotations publicly available, we utilize the open-source ImageCAS~\cite{ImageCASPaper} dataset, which contains 1000 CCTA scans with existing CA segmentations. The patients in the data set must be older than 18 years and with a documented medical history of ischemic stroke, transient ischemic attack and/or peripheral artery disease. 

The images consists of a mix of the 30\%-40\% phase or the 60\%-70\% phase depending on where the best CA images are obtained. The images have a size of $512 \times 512 \times (206-275)$ voxels, a planar resolution of $0.29-0.43\text{mm}^2$, and a spacing of $0.25-0.45$mm. The scanner used for the study does not have a full heart field-of-view and therefore the scans have been aquired over several rotations. This acquisition type sometimes induces \textit{step artefacts} with visible divisions between the scanfield from different rotations. 

\section{Methods}
\textbf{Initial whole heart segmentation.}
We use two models from TotalSegmentator v2.4.0~\cite{wasserthal2023totalsegmentator}: the \texttt{total} model (TSTOTAL), trained at 1.5mm voxel spacing, and the \texttt{heartchambers\_highres} model (TSHC), trained at submillimeter resolution. TSTOTAL segments 117 anatomical structures, while TSHC focuses on high-resolution cardiac anatomy, providing more accurate segmentations of the myocardium, LA, LV, RA, RV, aorta, and PA. Since TSHC does not segment the PVs or LAA, we use TSTOTAL to extract these structures, despite its lower resolution.

\textbf{LAA segmentaion using an implicit method.}
To reduce the computational complexity we use a two-step approach. 
Firstly, we estimate the center-of-mass of the LAA based on the crude LAA segmentation from TSTOTAL and crop the CCTA image with an isotropic bounding box with size $128^3$ voxels covering and a side length 70 mm. 

Secondly, we employ a state-of-the-art LAA segmentation tool to produce a refined LAA segmentation within the patch. 
We adopt the Neural Unsigned Distance Field (NUDF) method from \cite{NUDF} and use an implicit neural representation to parameterize the distance field of the LAA. 
In contrast to the standard segmentation methods which predict a voxelvise binary foreground/background mask, the implicit representation allows for representing detailed and continous structures with low computational demand. The NUDF method demonstrated the ability to segment the LAA with accuracies on the same order of magnitude as the voxel size in~\cite{NUDF}. For convenience, we represent the LAA labelmaps as closed surfaces and adapt the original NUDF code to use signed distance fields (SDFs), where all points inside the surface are associated with negative distances and points outside with positive distances.
Data preprocessing and training details are described in~\cite{NUDF}.

\textbf{Refinement of pulmonary vein segmentation.}
For anatomical plausibility, PVs must connect to the LA, a constraint often violated in TotalSegmentator labelmaps. To adress this, we developed a heuristic correction method:
\begin{enumerate}
    \item Create an SDF of both the PV labelmap and the LA labelmap,
    \item Sample HU values inside the LA and PV labelmaps and compute the 0.1\% and 99.9\% percentiles ($Q_\text{0.1}$ and $Q_\text{99.9}$)
    \item Identify the voxels where the following criteria are satisfied:
    \begin{enumerate}
        \item The distance to the LA is less than 7 mm.
        \item The distance to a PV is less than 7 mm.
        \item The HU value is between $Q_\text{0.1}$ and $Q_\text{99.9}$
    \end{enumerate}
\end{enumerate}
These voxels are added to the PV labelmap. A connected component analysis is then applied, keeping only components larger than 10,000 voxels. This process removes noise, enforces PV–LA connectivity, and matches the segmentation with the scan resolution.

\textbf{Label fusion.}
The final labelmap includes 10 distinct anatomical classes derived from four sources: heart chambers, myocardium, aorta and PA (from TSHC), refined PVs (from TSTOTAL), LAA (from NUDF), and refined CAs (from ImageCAS). Each voxel is assigned to a single class, as nested or hierarchical structures are not considered. Anatomical consistency is essential when merging labels. While CAs may course through the myocardium (myocardial bridging~\cite{lee2015myocardial}), they cannot intersect other organs. In practice, LAA segmentations occasionally include parts of the PVs, and CAs may border the LAA. The most coherent labelmaps are achieved using the following heuristic:  CAs > myocardium, PVs > LAA, LAA > CAs.

\section{Results}
\textbf{Anatomically coherent cardiac labels.}
The labelmaps are provided as compressed NIFTI files (.nii.gz) with integer voxel values, matching the original image dimensions and orientation. A thorough description of the labels can be found at https://github.com/Bjonze/Public-Cardiac-CT-Dataset \href{https://github.com/Bjonze/Public-Cardiac-CT-Dataset}{\faGithub}. 

\begin{figure}[h]
    \centering
    \includegraphics[width=\linewidth]{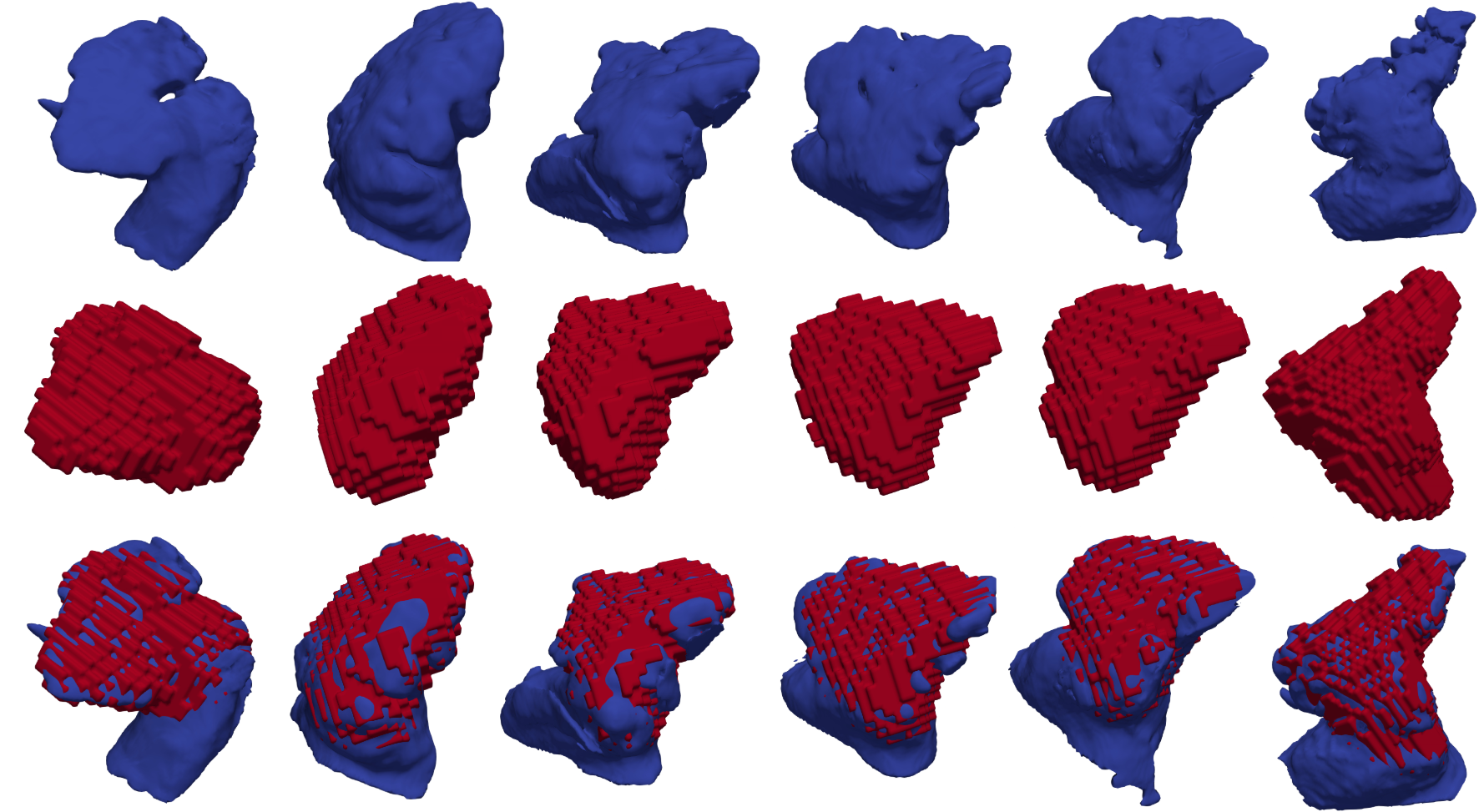}
    \caption{\textbf{Comparison of non-cropped, randomly selected scans from ImageCAS.} Top) Our segmentations. Middle) TSTOTAL segmentations. Bottom) combined segmentations. Shown are files 992, 957, 949, 572, 443, and 239 from ImageCAS. The voxelized, blocky nature of the TotalSegmentator LAA segmentations results from the low image resolution (1.5\,mm spacing) on which they were generated, combined with the use of marching cubes for rendering.
}
    \label{fig:NUDF_vs_TS}
\end{figure}

\textbf{LAA segmentation.}
Using the LAA segmentations produced by the lower resolution TSTOTAL model results in voxelized, blocky and occasionally incomplete segmentations. 
The LAA class from TSTOTAL only captures the largest morphological features of the LAAs, as illustrated in \textbf{Fig.~\ref{fig:NUDF_vs_TS}}. Consequently, these resulting surfaces are virtually impossible to classify according to the groups proposed by \cite{DIBIASE2012531} (chicken wings, cactus, cauliflower, windsock). 
In contrast, our segmentations, which uses an approximate spacing of 0.547\,mm, produce surfaces that are both smooth and accurately capture critical anatomical features such as trabeculation, lobes and finer structural details. These elements all play an important role in influencing blood flow and can create small pockets of blood stasis which can lead to stroke.

\begin{figure}[ht]
\includegraphics[width=0.49\textwidth]{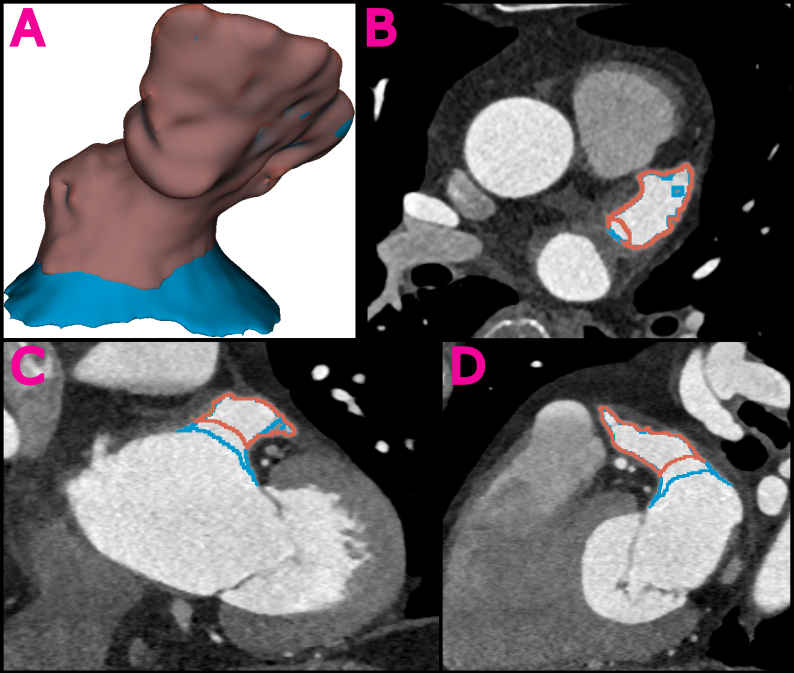}
\includegraphics[width=0.49\textwidth]{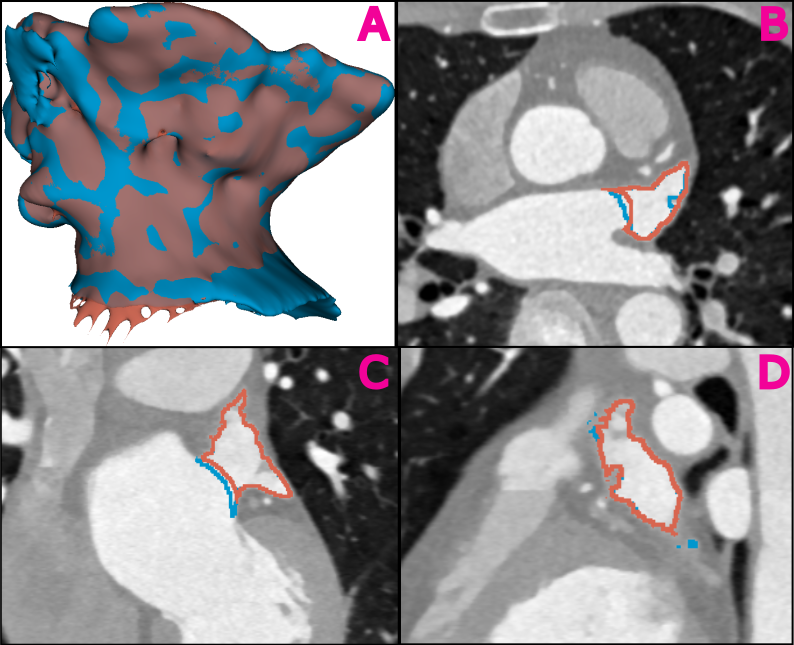}
\caption{\textbf{Expert annotation vs. NUDF segmentation.} Left) Lowest NUDF Dice score (81.5\%) on the 10 manually annotated cases from ImageCAS. Right) Highest NUDF Dice score (92.1\%). It is apparent that locating the ostium is challenging and contributes the most to loss of Dice score.} \label{fig:collected_segs}
\end{figure}

We additionally evaluated our model on 10 annotations created by a trained reader using the 3D Slicer segmentation tool. On these 10 segmentations, our model achieved a Dice score of 86.8\% $\pm 0.37$, indicating good performance despite the shift in CT image acquisition machines between our in-house dataset and ImageCAS. The segmentation with the best- and the worst Dice score are visualized in \textbf{Fig.~\ref{fig:collected_segs}}. 
We observe the largest deviations between the manual and our segmentations around the ostium. It is known to be anatomically ambigious to truly define the ostium, making this area particularly challenging \cite{leventic2019}.

\textbf{Refining the label map.}
When merging segmentations from different models, it’s important to consider how anatomical structures intersect. For continuous anatomies like the LA, PVs, and LAA, where no physical boundaries separate them, the segmentation should reflect their connected nature.

\begin{figure}[htbp]
\includegraphics[width=0.49\textwidth]{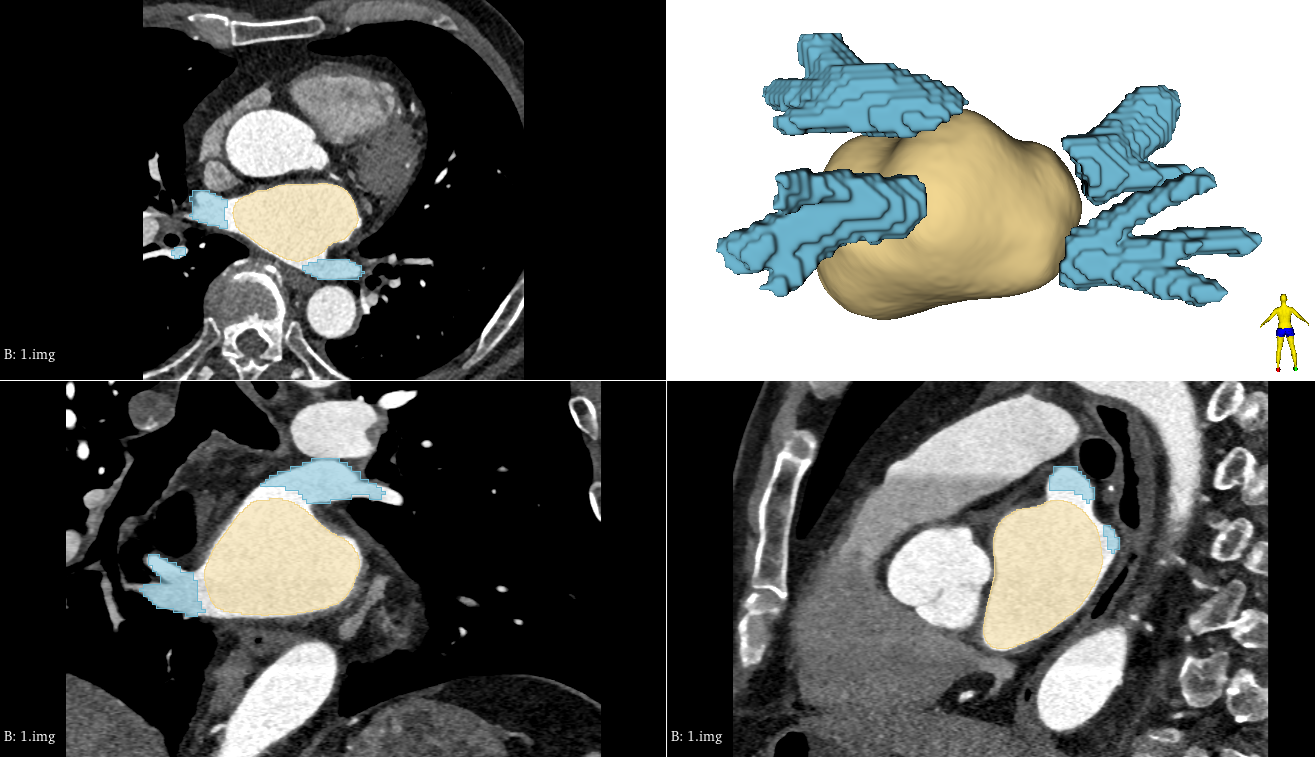}
\includegraphics[width=0.49\textwidth]{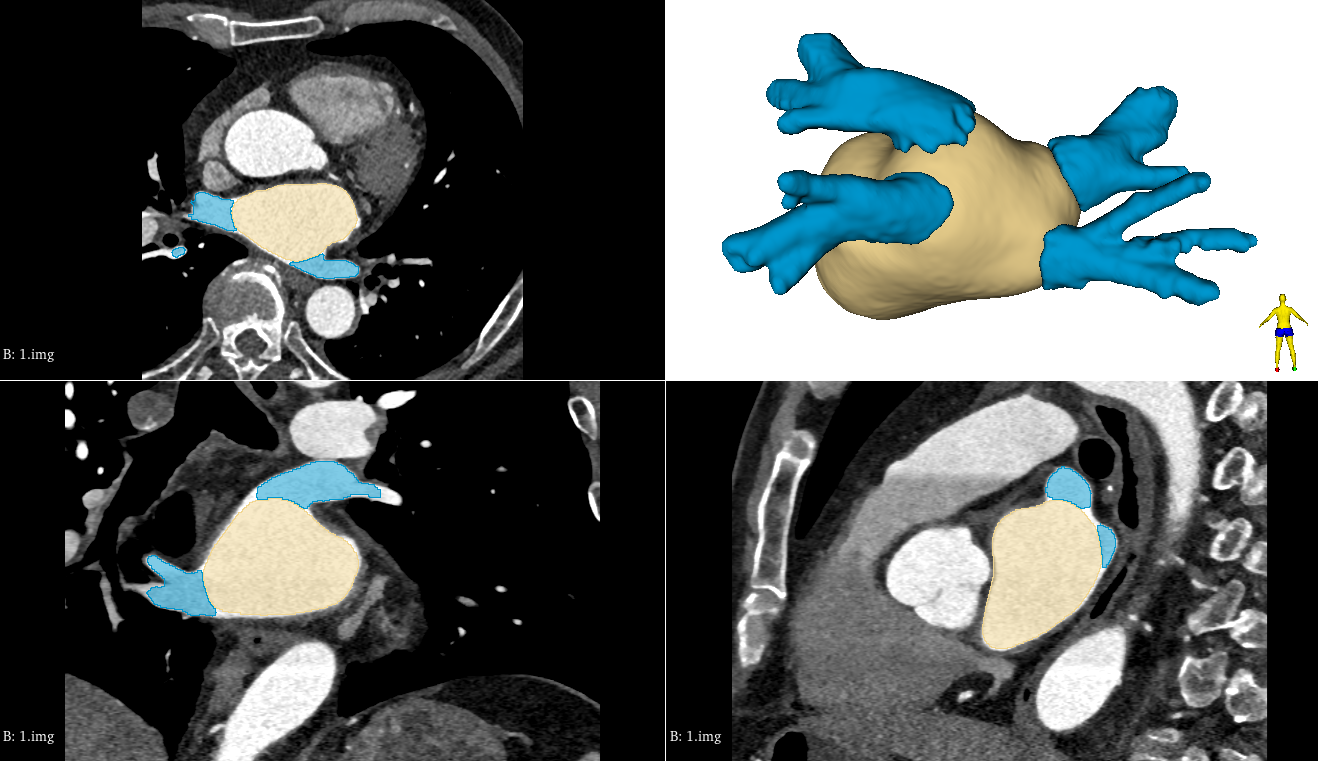}
\caption{\textbf{Comparison of TSTOTAL segmentations and postprocessing of pulmonary veins.} Left) LA with the original TotalSegmentator segmentation of the PVs. The blocky surface stems from the low resolution image the segmentation is created on. Right) LA with the refined PVs. Note the intensity change of the contrast enhanced blood pool due to step artefacts caused by the scanner acquisition.} \label{fig:RefinedPV}
\end{figure}

\textbf{Fig.~\ref{fig:RefinedPV}} illustrates this by comparing raw PV segmentations with the refined labelmap. The refined PVs are smoother and fully connected to the LA, whereas the unprocessed TSTOTAL output shows a visible gap, likely due to LA undersegmentation in TSHC.

\textbf{Dataset quality control.}
ImageCAS contains scans of varying quality, which affects the accuracy of the provided segmentations. Additionally, the raw ImageCAS data can exhibit strong step artefacts and occasional patient movement. This results in volumes divided into three parts that are stitched together in a post-acquisition step and varying contrast in the same scan, causing issues for a direct HU analysis. 
We further observe that the upper part of the heart is outside the scanners field of view in a significant amount of the images in ImageCAS. 
Since the LAA is extruding on the upper part of the heart, we include a check for assessing whether the LAA is cut by the field of view and find 330 images where the LAA segmentation touches the boundary of the image. 

Alongside the segmentations, we provide a list of observed scan issues at https://github.com/Bjonze/Public-Cardiac-CT-Dataset \href{https://github.com/Bjonze/Public-Cardiac-CT-Dataset} {\faGithub}. 
Whether the issues warrant an exclusion of the data in a downstream analysis task will depend on the problem, and we leave this decision to the user.

\section{Discussion and Conclusion} 
Despite TotalSegmentator’s impact on medical image segmentation, important structures like the LAA, PVs, and CAs often lack the segmentation quality seen in larger organs. This paper introduces the first open-source dataset of high-resolution, anatomically consistent labelmaps for these cardiac structures. Alongside the labelmaps, we include a list of LAAs that are outside the field of view of the scanner, contains step artefacts, or other types of defects, making it easier for future users to navigate the data. The released dataset provide a freely available resource for high-quality LAA, PV, and CA anatomies to be used in future exploration of anatomical variability of complex heart structures.

Note that the dataset includes a mix of ES and ED scans without clear labeling of the cardiac phase, which may affect analyses like volume estimation. Additionally, no patient metadata (e.g., age, sex, disease status) is available.

Despite artefacts and out-of-view issues, the ImageCAS dataset remains significantly larger than other open-source CCTA datasets. With the addition of detailed LAA segmentations, it becomes the first large-scale, open-source LAA database. Previous efforts to analyze LAA shape and morphology have struggled to reach consensus, partly due to the limited size and private nature of existing datasets. By offering a large-scale, publicly available cardiac dataset, we aim to encourage research into robust characterization of LAA morphology and enable more meaningful comparisons between methods. 

We demonstrate that high quality coherent labelmaps can be obtained from a set of unlabelled CCTA images using a combination of TotalSegmentator outputs, neural distance field based LAA segmentation and label refinement methods. 
We demonstrate how this can create a segmentation map of anatomical coherent heart structures on 1000 CCTA images from the open source ImageCAS dataset.
In addition to what can be achieved with standard TotalSegmentator outputs, our detailed LAA segmentations offer the possibility of analysing the morphology of 670 LAA shapes in coherence with LA and PV segmentations. 
By making these resources widely available, we aim to support the ongoing research in stroke etiology and cardiovascular disease stratification.

\bibliographystyle{splncs04}
\bibliography{references.bib}
\end{document}